# Can Transfer Neuroevolution Tractably Solve Your Differential Equations?

Jian Cheng Wong, Abhishek Gupta, and Yew-Soon Ong, *Fellow, IEEE*

*Abstract*— This paper introduces neuroevolution for solving differential equations. The solution is obtained through optimizing a deep neural network whose loss function is defined by the residual terms from the differential equations. Recent studies have focused on learning such *physics-informed neural networks* through stochastic gradient descent (SGD) variants, yet they face the difficulty of obtaining an accurate solution due to optimization challenges. In the context of solving differential equations, we are faced with the problem of finding globally optimum parameters of the network, instead of being concerned with out-of-sample generalization. SGD, which searches along a single gradient direction, is prone to become trapped in local optima, so it may not be the best approach here. In contrast, neuroevolution carries out a parallel exploration of diverse solutions with the goal of circumventing local optima. It could potentially find more accurate solutions with better optimized neural networks. However, neuroevolution can be slow, raising tractability issues in practice. With that in mind, a novel and computationally efficient *transfer neuroevolution* algorithm is proposed in this paper. Our method is capable of exploiting relevant experiential priors when solving a new problem, with adaptation to protect against the risk of negative transfer. The algorithm is applied on a variety of differential equations to empirically demonstrate that transfer neuroevolution can indeed achieve better accuracy and faster convergence than SGD. The experimental outcomes thus establish transfer neuroevolution as a noteworthy approach for solving differential equations, one that has never been studied in the past. Our work expands the resource of available algorithms for optimizing physics-informed neural networks.

## I. Introduction

Solving ordinary differential equations (ODEs) and partial differential equations (PDEs) is the cornerstone of scientific modelling in modern science and engineering [1-3]. The solution of these differential equations allows us to understand and make predictions on the behaviors of physical systems in a wide variety of scientific problems, including heat and mass transfer, optics, acoustics, material elasticity, electromagnetic waves, fluid dynamics, and many other dynamical processes in physics [4, 5], sociology [6], finance and economics [7]. The idea of using neural networks to solve differential equations goes back to early works in the 1990s [8-14]. Given the surging popularity of deep learning lately and advancement in computing capability, there has been renewed interest in designing deep neural networks (DNNs) to solve differential equations [15-23]. In brief, a DNN—also termed herein as a physics-informed neural network (PINN) as per Raissi *et al.* [24]—is constructed such that its output $\hat{u}(x, t)$ generates the solution $u$ of the governing differential equations in space $x \in \Omega$ and time $t \in [0, T]$ domains. The loss function of such a PINN is defined by the residual terms from the differential equations, under prescribed initial and boundary conditions. In principle, *infinite* training data instances are accessible by sampling arbitrary points within a problem's spatial-temporal domain. The loss thus acts as a penalty to constrain the PINN from violating the governing equations at *all* points. *In other words, the problem of solving differential equations is transformed to one of global optimization, in which the PINN loss is to be minimized to zero.*

Different from classical numerical solvers [25], the PINN approach has the main advantage of being mesh-free. Meshing (i.e., spatial discretization) itself is a nontrivial task. Classical numerical schemes are prone to failure in finding the right solution if the meshing is not appropriately done. Their accuracy is limited by the size of the discretization and interpolation scheme used (usually linear). Being universal approximators, PINNs offer superior approximation or even exact replication to the solution. Moreover, the solution via PINN is differentiable. Its relatively compact representation requires lower memory demand for storage. Beyond solving differential equations (referred to as the forward problem), physics-informed neural networks can be further extended to solve inverse problems such as designing metamaterials [26], inferring unknown dynamic from observations [27, 28], and quantifying fluid flows from visualizations or sensors data [29-31]; see examples in Figure 1. They offer a path to *rationalizable* artificial and computational intelligence systems that are consistent with fundamental physics laws [32]. These advantages brought by synergizing PINNs and differential equations are however attainable at the cost of a steep optimization challenge caused by high-dimensional, non-convex loss function landscapes.

When considering the optimization of physics-informed neural networks, there has been a lack of research investigation

Jian Cheng Wong, Agency for Science, Technology and Research (A*STAR) and Nanyang Technological University, Singapore.

Abhishek Gupta, Agency for Science, Technology and Research (A*STAR), Singapore.

Yew-Soon Ong, Agency for Science, Technology and Research (A*STAR) and Nanyang Technological University, Singapore.

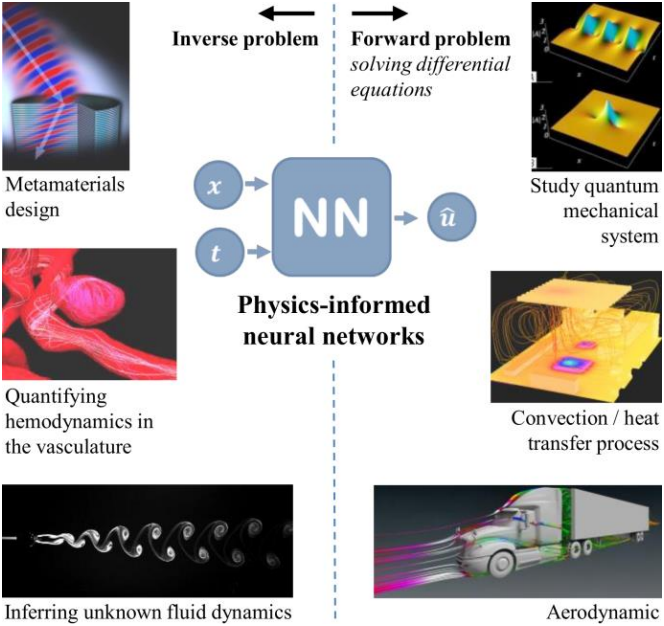

Fig. 1. Applications of physics-informed neural networks. By solving differential equations, accurate predictions on the behaviors of physical systems can be made in a wide variety of problems across scientific fields. PINNs can also be extended to solve inverse problems.

beyond mainstream gradient descent methods, such as stochastic gradient descent (SGD) and Broyden–Fletcher–Goldfarb–Shanno (BFGS) algorithms. Their known tendency of becoming trapped in local minima can nevertheless hinder a PINN from approximating the right solution. Although SGD has been very successful in deep learning, the emphasis is typically to learn a model from noisy data with good out-of-sample generalization. Over-fitting to the training data is discouraged, and a local minimum often suffices in achieving high test accuracy. However, the requirement is somewhat different when it comes to solving differential equations, in which globally optimum model parameters are sought. Generalizability and over-fitting are not a concern given access to potentially infinite training data covering the entire problem domain; rather, having a pre-maturely converged physics-informed neural network implies an unphysical solution. For these reasons, it is contended that SGD may not necessarily be the best approach for optimization in this domain.

An alternative approach comes from the field of *neuroevolution* [33], which uses evolutionary algorithms (EAs) to optimize PINNs. Their main conceptual distinction from SGD is that SGD follows a single gradient direction, whereas EAs search with a population of diverse solutions with the goal of circumventing local optima [34]. This makes neuroevolution a different paradigm from gradient descent, since the notion of diversity does not explicitly exist in the latter [35]. As such, neuroevolution offers a promising substitute to SGD for global optimization [36]; such as for solving differential equations. In particular, it is demonstrated in this paper that neuroevolution via a state-of-the-art EA, namely *natural evolution strategies* (NES) [37], outperforms SGD in solving a variety of differential equations, providing physically accurate solutions. Neuroevolution is thus highlighted as a noteworthy approach for solving differential equations.

Neuroevolution can however be slow to converge compared to gradient descent. To enhance computational tractability, we propose a novel augmentation of neuroevolution with *transfer optimization* [38-40], where information in the form of *experiential priors* are reused from past (*source*) problem instances to boost the *target* search. In a scientific study, it is common for a single set of differential equations to be evaluated under different environments and boundary conditions. Relevant experiences are therefore naturally accumulated over time, and can be exploited given any new target problem. To this end, we develop a new transfer optimization method which can be integrated with probabilistic model-based evolution strategies, such as NES (or even others like CMA-ES or OpenAI-ES [41, 42]). Different from the commonly used transfer strategy of fixing pre-trained neural network layers [43, 44], our method features a probability mixture model-based adaptive design to protect the algorithm from the risk of negative transfer. The mixture model allows the joint processing of multiple sources, adaptively selecting the one that is most relevant to the target task. In the experimental study, the proposed transfer optimization algorithm is integrated with the NES, empirically demonstrating improvements in convergence speed and accuracy over baseline neuroevolution and SGD.

The remainder of the paper is organized as follows. In the next section, physics-informed neural networks for differential equation problem is described. In Section III, the methodology of neuroevolution via probabilistic model-based evolution strategies is introduced. The proposed method for achieving transfer neuroevolution is then presented in Section IV. Section V presents the experimental study to illustrate the competitive advantage of transfer neuroevolution over SGD across several test problems. Finally, Section VI contains the conclusion and directions for future research.

## II. PROBLEM SETUP: PHYSICS-INFORMED NEURAL NETWORKS FOR DIFFERENTIAL EQUATIONS

### A. Differential Equations

Consider differential equations of the general form:

$$u_t(x, t) + \mathcal{N}_x[u(x, t)] = 0, \quad x \epsilon \Omega, t \epsilon [0, T], \quad (1)$$

$$u(x, 0) = u_o(x), \quad x \epsilon \Omega, \quad (1b)$$

$$\mathcal{B}[u(x, t)] = g(x, t), \quad x \epsilon \partial\Omega, t \epsilon [0, T], \quad (1c)$$

where $u_t(x, t)$ is the temporal derivative, and $\mathcal{N}_x[\cdot]$ is a general nonlinear differential operator which includes non-linear terms of spatial derivatives, such as the first and second order derivatives $u_x(x, t)$ and $u_{xx}(x, t)$, respectively. The differential equation (1) usually describes certain dynamical processes in the physical world. The corresponding spatial domain can be of 1-, 2- or 3-dimensions. As an example, the Burgers' equation reads $u_t + u \cdot u_x - \nu \cdot u_{xx} = 0$, by having $\mathcal{N}_x[u] = u \cdot u_x - \nu \cdot u_{xx}$. The nonlinear differential operator in Burgers' equation describes the advection and diffusion processes of the physical quantity $u$ in 1D, and contains a problem specific diffusion coefficient $\nu$.

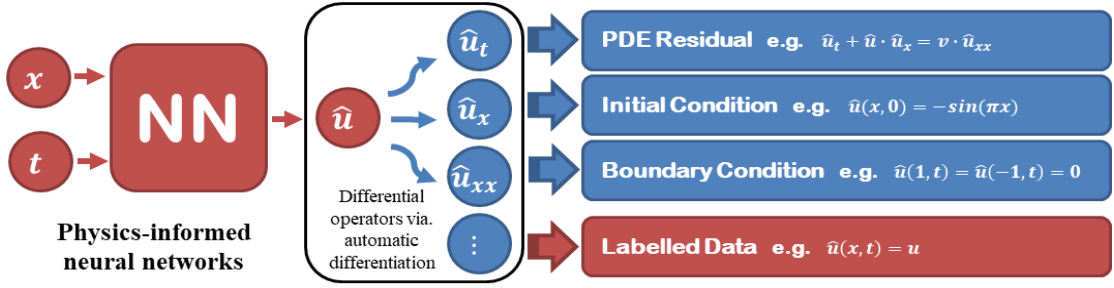

Fig. 2. Physics-informed neural networks consist of additional residual (loss) terms from the differential equation, initial and boundary conditions (blue), which require the computation of differential operators of the neural network output with respect to the inputs. It can emulate the solution of differential equations without the need for labelled data. We have a global optimization problem (instead of a learning problem) to find the most physically accurate solution.

The interest of this paper lies in finding the solution $u(x, t)$ which satisfies the differential equation (1) across space $x \epsilon \Omega$ and time $t \epsilon [0, T]$ domains. Such a solution may not be unique unless sufficient initial condition (1b) and/or boundary condition (1c) are given. The initial condition at $t = 0$ is defined by $u_o(x)$, and the boundary operator $\mathcal{B}[\cdot]$ enforces the desired condition $g(x, t)$ at the domain boundary $\partial\Omega$. This $\mathcal{B}[\cdot]$ can be an identity operator (Dirichlet boundary condition) or a differential operator (Neumann boundary condition). A meaningful solution would need to satisfy all these conditions, in addition to the main differential equation.

### B. Physics-Informed Neural Networks

To solve the differential equation in (1), the neural network approach constructs a PINN representation $\hat{u}(x, t; \boldsymbol{w})$ to emulate the unknown solution $u$, with network parameters $\boldsymbol{w} = \{w_i\}_{i=1}^d$ to be optimized. In the present study, the network architecture and other hyper-parameters are specified, making $\boldsymbol{w}$ refer to the weights of the neural network. Such a neural network is termed as "physics-informed neural network", because it uses the residual terms from the differential equation (1), and the prescribed initial (1b) and boundary (1c) conditions as the loss function, which is defined below:

$$\mathcal{L} = \mathcal{L}_{DE} + \beta_{IC} \cdot \mathcal{L}_{IC} + \beta_{BC} \cdot \mathcal{L}_{BC}, \qquad (2)$$

where,

$$\mathcal{L}_{DE} = \|\hat{u}_t(\cdot\,; \boldsymbol{w}) + \mathcal{N}_x[\hat{u}(\cdot\,; \boldsymbol{w})]\|_{\Omega \times [0,T]}^2, \qquad (2b)$$
$$\mathcal{L}_{IC} = \|\hat{u}(\cdot\,, 0\,; \boldsymbol{w}) - u_0\|_{\Omega}^2, \qquad (2c)$$
$$\mathcal{L}_{BC} = \|\mathcal{B}[\hat{u}(\cdot\,; \boldsymbol{w})] - g(\cdot)\|_{\partial\Omega \times [0,T]}^2. \qquad (2d)$$

Here, $\|h(y)\|_y^2 = \int_y |h(y)|^2 \, dy$ where $y \epsilon \mathcal{Y}$. The defined integral loss (2) has a global minimum of zero, given that the differential equation and the initial and boundary conditions are exactly satisfied everywhere in the problem domain $\Omega \times [0, T]$. At $\mathcal{L} = 0$, the PINN output exactly emulates the *unknown* true solution of the differential equation. In this regard, we are faced with a global optimization problem (instead of a learning generalization problem) for finding the most physically accurate solution to the target differential equation.

The relative weights $\beta s$ in (2) control the trade-off between different terms in the loss function during the optimization process, and may need to be scaled in a problem-specific way.

PINNs may also include conventional data loss (for example, when solving an inverse problem). In the case of solving ODEs and PDEs, the loss function only comprises of the residual terms with respect to the differential equation, initial and boundary conditions; see illustration in Figure 2.

### C. Computation of the Loss

The computation of the loss (2) involves substitution of the PINN output $\hat{u}$ into the differential equation for evaluating the residuals (2b) over the computational domain, as well as matching the output $\hat{u}$ against initial condition (2c) at $t = 0$, and boundary conditions (2d) over the domain boundary $\partial\Omega$. For certain types of activation functions (for example, sigmoid, softplus, and tanh), the PINN output is higher order differentiable with respect to its spatial and temporal inputs. These differential operators, such as $\hat{u}_t(x, t; \boldsymbol{w})$, $\hat{u}_x(x, t; \boldsymbol{w})$, $\hat{u}_{xx}(x, t; \boldsymbol{w})$, are required for the evaluation of the loss. They can be conveniently obtained via automatic differentiation [45].

Although the integral loss terms (2b-d) are defined over a continuous computational domain, for practical reasons, during the loss evaluation we compute the mean squared residuals over a set of $m$ spatial-temporal collocation points $\mathcal{D} = \{(x_i, t_i)\}_{i=1}^m$. These points are sampled (for example, using randomized Latin hypercube sampling) from the respective computational domains. In this paper, a dynamic sampling scheme is adopted such that for every loss evaluation a new batch of $m$ collocation points are generated. The evaluated loss provides us an *approximate estimation* on how well the differential equation and the prescribed initial and boundary conditions are being satisfied.

## III. METHODOLOGY: NEUROEVOLUTION

The central thesis of this paper is the untapped efficacy of evolutionary algorithms for optimizing the PINN in generating good solutions to the differential equations. In particular, the objective is to find the best $d$-dimensional neural network weights $\boldsymbol{w} \in R^d$ which minimize the loss (2). As opposed to typical gradient descent methods such as SGD, which searches along a single gradient direction, EAs search with a population of diverse solutions $\{\boldsymbol{w}_k\}_{k=1}^\lambda$ for better *fitness*. Here, $\lambda$ is the population size and the fitness $f = -\mathcal{L}$ refers to the negative loss of the physics-informed neural network, which is evaluated on a dynamically sampled batch of $m$ collocation points. The diversity in an EA population could potentially be the key factor to overcome local optima and thus achieve better optimized neural networks.

Among EAs, there is a subgroup of techniques such as NES, CMA-ES, and OpenAI-ES, that adopt a probabilistic model, called the *search distribution*, to represent the population. In a nutshell, they keep tracing an evolving search distribution and produce pseudo-offspring by drawing new samples from the distribution. As the search progresses, the distribution is iteratively updated towards regions with higher population fitness. All probabilistic model-based evolution strategies follow this basic principle, although they may differ in their distribution update mechanism. They have shown to be an effective method to evolve neural network weights $w$ in continuous domains, by using a $d$-dimensional multivariate normal search distribution [37, 42, 46-49].

For demonstration purposes, we consider the state-of-the-art NES as the baseline neuroevolution algorithm in this paper. The underlying objective function of NES can be expressed as:

$$\mathcal{J}(\boldsymbol{\theta}) = \mathrm{E}_{\boldsymbol{\theta}}[f(\boldsymbol{w})] = \int f(\boldsymbol{w})\, \pi(\boldsymbol{w}|\boldsymbol{\theta})\, d\boldsymbol{w}, \qquad (3)$$

which is the expected fitness of the population represented by probabilistic model $\pi(\boldsymbol{w}|\boldsymbol{\theta})$, where the search distribution is specified by *distributional parameters* $\boldsymbol{\theta}$. In this paper, the $d$-dimensional multivariate normal search distribution is parameterized by $\boldsymbol{\theta} = (\mu, A)$, where $A A^T = \Sigma$. The distribution mean $\mu \in R^d$ and full covariance matrix $\Sigma \in R^{d \times d}$ characterize the search center and mutation. A candidate solution for the neural network weights is therefore a realization of the normally distributed random variable $\boldsymbol{w} \sim N(\mu, \Sigma)$.

It is worth noting that NES searches in the space of distributional parameters, but not the problem space. In contrast, the objective function for typical gradient descent methods such as SGD is to search for an optimal $\boldsymbol{w}$ directly; i.e., $\mathcal{J}(\boldsymbol{w}) = \mathcal{L}(\boldsymbol{w})$. This further highlights the conceptual distinction between the two approaches.

In particular, NES utilizes the *search gradient* $\nabla_{\boldsymbol{\theta}} \mathcal{J}(\boldsymbol{\theta})$ on the expected fitness (3) to evolve the search distribution. The gradient can be estimated from a population of *pseudo-offspring* drawn from the current search distribution as [37]:

$$\nabla_{\boldsymbol{\theta}} \mathcal{J}(\boldsymbol{\theta}) \approx \frac{1}{\lambda} \sum_{k=1}^{\lambda} f(\boldsymbol{w}_k)\, \nabla_{\boldsymbol{\theta}} \log \pi(\boldsymbol{w}_k|\boldsymbol{\theta})), \qquad (4)$$

where $\lambda$ is the population size and $f(\boldsymbol{w}_k)$ is the fitness of the $k^{\text{th}}$ sampled offspring. Then, the distributional parameters are updated based on the estimated search gradient as:

$$\boldsymbol{\theta} \leftarrow \boldsymbol{\theta} + \eta \cdot \mathrm{F}^{-1} \nabla_{\boldsymbol{\theta}} \mathcal{J}(\boldsymbol{\theta}), \qquad (5)$$

where $\eta$ is the learning rate. In (5) the search gradient is normalized by the inverse of the *Fisher information matrix* $\mathrm{F} = \mathrm{E}_{\boldsymbol{\theta}}[\nabla_{\boldsymbol{\theta}} \log \pi(\boldsymbol{w}|\boldsymbol{\theta}) \nabla_{\boldsymbol{\theta}} \log \pi(\boldsymbol{w}|\boldsymbol{\theta})^{\mathrm{T}}]$ (i.e., the variance of the gradient) for the given distributional parameters to form the *natural gradient* $\widetilde{\nabla_{\boldsymbol{\theta}}} \mathcal{J}(\boldsymbol{\theta}) = \mathrm{F}^{-1} \nabla_{\boldsymbol{\theta}} \mathcal{J}(\boldsymbol{\theta})$, which is the hallmark of NES. This natural gradient takes into consideration the uncertainty of gradient estimates when providing an ascent direction in the space of distributional parameters.

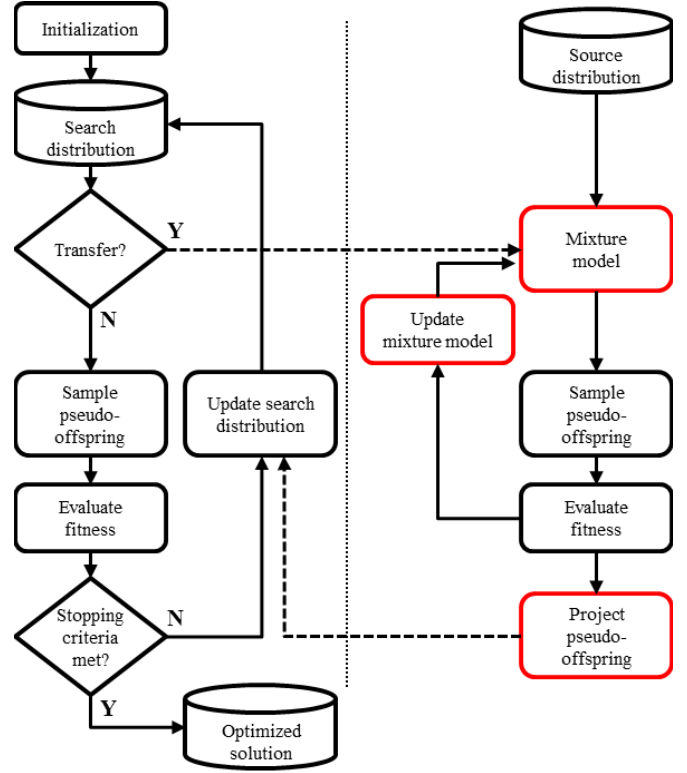

Fig. 3. A conceptual illustration of the proposed mixture model-based adaptive transfer, featuring the methodology (1) to construct and update a mixture model of the target search and source distributions, and (2) to influence the search distribution update on the target problem based on the projection of pseudo-offspring sampled from the source distributions (boxes highlighted in red). The proposed transfer method is suitable for general probabilistic model-based evolution strategies as shown on the left side of the vertical dotted line, with their interface indicated by dashed arrows.

## IV. TRANSFER NEUROEVOLUTION

In this section, we augment neuroevolution with transfer optimization, boosting search efficiency by reusing experiential priors from related source problem instances. This capability is enabled via a mixture model-based adaptive transfer method. The proposed method is suitable for general probabilistic model-based evolutionary strategies. It is worth noting that the approach makes no strict assumption on the synergy between source and target problems. The adaptive transfer method features a dynamic mechanism to automatically exploit useful experience from source problems, such that better quality pseudo-offspring can be induced to effectively influence the target search distribution update. Importantly, the adaptation mechanism is able to retreat from irrelevant sources, to curb negative transfer. Figure 3 gives the conceptual illustration for the proposed mixture model-based adaptive transfer method.

### A. Transfer via Mixture Modelling

Let us consider a search distribution $\pi(\boldsymbol{w}|\boldsymbol{\theta})$ for the target optimization problem, and a single source distribution $\varphi(\boldsymbol{w})$ for simplicity of exposition. It is assumed that the source is expressed in the form of a distributional prior, for example, a search distribution acquired from successful evolution of the same neural network with NES (or any other probabilistic model-based evolution strategies like CMA-ES or OpenAI-ES) for solving a similar differential equations problem. To facilitate knowledge transfer, a mixture model to unify the target and source distributions is defined as:

$$m(\boldsymbol{w}) = \alpha_\pi \, \pi(\boldsymbol{w}) + \alpha_\varphi \, \varphi(\boldsymbol{w}), \tag{6}$$

where $\alpha_\pi$ and $\alpha_\varphi$ are the mixing coefficients for the target and source distribution components, respectively. This mixture model can be initiated with arbitrary mixing coefficients, subject to the constraint $\alpha_\pi + \alpha_\varphi = 1$. In each iteration of neuroevolution, a predefined parameter is used to ascertain if the knowledge transfer mode is to be *activated*. In the transfer mode, $\lambda$ pseudo-offspring are sampled from the mixture model $m(\boldsymbol{w})$ instead of the search distribution $\pi(\boldsymbol{w}|\boldsymbol{\theta})$, in effect inducing pseudo-offspring from the source to the target problem. Otherwise, the algorithm continues to draw pseudo-offspring from the target search distribution.

It is important to note that the target search distribution is *evolving* (see Section IV-C), while the source distribution is *fixed* during the optimization process. The key step is how to dynamically update the mixing coefficients $\alpha_\pi$ and $\alpha_\varphi$ when the search progresses, where $\alpha_\varphi$ can be viewed as the degree of transfer from source to target. If the source problem is beneficial, we would want to increase $\alpha_\varphi$ so that more high-quality pseudo-offspring can be sampled from the source distribution to influence the target search. However, after a certain point, the pseudo-offspring from the source may no longer remain competitive with those sampled from the evolved target search distribution. At that point, we would want to gradually reduce $\alpha_\varphi$, eventually deactivating the source distribution from the mixture model, i.e., setting $\alpha_\varphi = 0$.

Decidedly, how beneficial a pseudo-offspring is to the target problem shall be reflected by its fitness. To this end, we propose to update the mixing coefficients towards better *expected fitness* under the mixture model, with the following generalization of (3):

$$\mathcal{J}^m = \int f(\boldsymbol{w}) \left[ \alpha_\pi \, \pi(\boldsymbol{w}) + \alpha_\varphi \, \varphi(\boldsymbol{w}) \right] d\boldsymbol{w}. \tag{7}$$

Then, the gradient estimates for $\alpha_\pi$ and $\alpha_\varphi$ on the expected fitness can be derived, using the log-likelihood trick, as:

$$\nabla_{\alpha_\pi} \mathcal{J}^m = \sum_{k=1}^{\lambda} f(\boldsymbol{w}_k) \, \nabla_{\alpha_\pi} \log \left( \alpha_\pi \, \pi(\boldsymbol{w}_k) + \alpha_\varphi \, \varphi(\boldsymbol{w}_k) \right)$$
$$= \sum_{k=1}^{\lambda} f(\boldsymbol{w}_k) \, \frac{\pi(\boldsymbol{w}_k)}{\alpha_\pi \, \pi(\boldsymbol{w}_k) + \alpha_\varphi \, \varphi(\boldsymbol{w}_k)}, \tag{8a}$$

and,

$$\nabla_{\alpha_\varphi} \mathcal{J}^m = \sum_{k=1}^{\lambda} f(\boldsymbol{w}_k) \, \frac{\varphi(\boldsymbol{w}_k)}{\alpha_\pi \, \pi(\boldsymbol{w}_k) + \alpha_\varphi \, \varphi(\boldsymbol{w}_k)}, \tag{8b}$$

given $\lambda$ pseudo-offspring $\boldsymbol{w}_1, \boldsymbol{w}_2, ..., \boldsymbol{w}_\lambda$ sampled from the mixture model and their fitness $f(\boldsymbol{w}_1), f(\boldsymbol{w}_2), ..., f(\boldsymbol{w}_\lambda)$. With these gradient estimates, the mixing coefficients can be updated as follows:

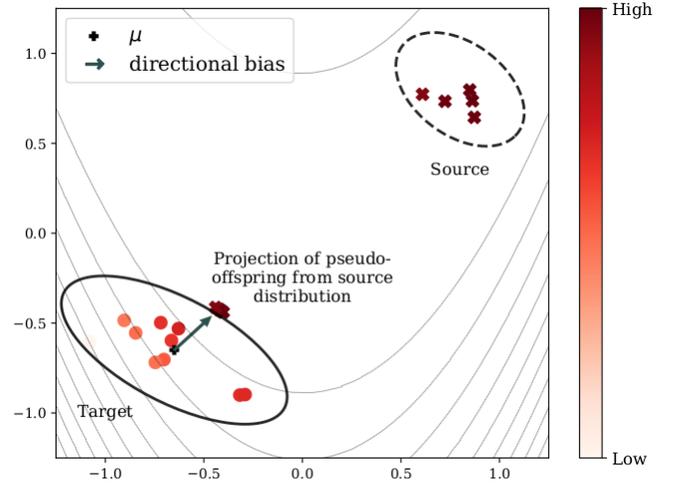

Fig. 4. An illustration of projecting the pseudo-offspring (marker: cross) produced by the source distribution towards the target search distribution. The projected pseudo-offspring are within $r$ Mahalanobis distance from the search distribution center, while still preserving the same directional and fitness information (indicated by color scale). By combining them with the pseudo-offspring originating from the target distribution (marker: round), a definite directional bias is induced on the evolution of the target search without absurdly changing it in a single update, thus promoting numerical stability.

$$\alpha_\pi \leftarrow \alpha_\pi + \eta_\alpha \cdot \nabla_{\alpha_\pi} \mathcal{J}^m, \tag{9a}$$
$$\alpha_\varphi \leftarrow \alpha_\varphi + \eta_\alpha \cdot \nabla_{\alpha_\varphi} \mathcal{J}^m, \tag{9b}$$

where $\eta_\alpha$ is the learning rate. Note that the constraint $\alpha_\pi + \alpha_\varphi = 1$ can be easily imposed by normalization. The mixture model formulation described above can be seamlessly extended to multiple components for multiple sources transfer, making the method more powerful since the chance of having useful experience included increases with the number of sources.

### B. Influencing Evolution of the Target Search Distribution

After sampling pseudo-offspring from the mixture model and evaluating their fitness, the target search distribution is to be updated. Broadly speaking, we expect the target distribution to gradually evolve with the guidance of pseudo-offspring transferred from the source. In this way, it has a greater chance to explore diverse and high-quality solutions along the search path. However, absurdly large changes to the mean $\mu$ of the target search distribution in a single update step should be avoided. At the same time, it is not advisable to be overly greedy in expanding the covariance matrix $\Sigma$ when good pseudo-offspring (from the source) are very far away from the target distribution. A direct induction of pseudo-offspring from the source may thus result in such deleterious outcomes, often leading to numerical instability issues.

To understand this issue, let us assume a pseudo-offspring $\boldsymbol{w}_k$ with fitness $f(\boldsymbol{w}_k)$ being induced from the source distribution. Its contribution to the distributional parameter gradient estimates are $f(\boldsymbol{w}_k)\boldsymbol{z}_k$ and $f(\boldsymbol{w}_k)(\boldsymbol{z}_k\boldsymbol{z}_k^T - \mathbb{I})$ [49, 50], where $\boldsymbol{z}_k = A^{-1}(\boldsymbol{w}_k - \mu)$ maps $\boldsymbol{w}_k$ into the target search distribution's natural coordinates; $\mu$ and $A$ are the center and the square root of the covariance matrix of the target search distribution, respectively. If $\boldsymbol{w}_k$ is far from $\mu$ in relation to $A$, then the value $\boldsymbol{z}_k = A^{-1}(\boldsymbol{w}_k - \mu)$ will be extremely large. As a result, $z_k z_k^T$ will be further exaggerated. This could cause the gradient estimates to (numerically) explode, causing failure of the distributional parameter updates. Such instability is likely to occur even in moderate dimensions, due to distribution sparsity as a consequence of the curse of dimensionality.

To overcome the instability issue, we propose to project the source distribution's pseudo-offspring closer to the target search distribution before updating the distributional parameters. While doing so, however, the original fitness values of the projected offspring are retained. A simple strategy to perform such projection, under the assumption that the target follows a multivariate normal distribution—a common practice in most probabilistic model-based evolution strategies—is outlined below. First, we define a tunable parameter $r$ to represent a threshold Mahalanobis distance from the distribution center. Choosing, say, $r=3$ implies that only those pseudo-offspring that are more than 3 Mahalanobis distance away from the target distribution center are projected. In particular, the direction vector $d_k = w_k - \mu$ is computed, and then the corresponding pseudo-offspring $w_k$ is projected to $\widetilde{w}_k$ as shown below (so that it lies within $r$ Mahalanobis distance from the center):

$$w_k \to \widetilde{w}_k = \mu + d_k \times \min\left(1, \frac{r}{\|A^{-1}d_k\|}\right). \tag{10}$$

Note that the denominator $\|A^{-1}d_k\|$ is the Mahalanobis distance between $w_k$ and the target distribution center. As demonstrated in Figure 4, these projected pseudo-offspring are now closer to the target distribution, while preserving directional and fitness information for the update. As a result, they induce a definite directional bias on the evolution of the target search distribution without absurdly changing it in a single update.

Our formulation of transfer neuroevolution with mixture modelling and pseudo-offspring projection is generic for probabilistic model-based search, and does not depend on a specific algorithm. The next section presents a particular instantiation of this method by implementing it on a variant of NES. This transfer NES algorithm will be subsequently applied to solve differential equations problems.

### C. An Instantiation: Transfer NES (tNES) Algorithm

In this section, transfer neuroevolution with a variant of NES, namely the *exponential NES* (xNES) algorithm [50], is presented. The xNES efficiently updates the mean $\mu$ and the full covariance matrix $\Sigma = A\,A^T$ of a search distribution via a coordinate transformation trick, to avoid trivial operations on computing and inverting the Fisher information matrix in (5). The details of the xNES algorithm can be found in [50]. Algorithm 1 presents the pseudocode of our transfer implementation on xNES, which is hereafter referred to as *t*NES.

The *t*NES algorithm requires the following inputs: fitness function, initial search distribution, a source distribution, and initial mixing coefficients. In the iteration loop, the *t*NES algorithm firstly checks for the transfer flag. If the transfer mode is not activated, $\lambda$ pseudo-offspring are sampled from the

---

**Algorithm 1:** Pseudocode of *t*NES

**Input:** optimization problem $f \in R^d \to R$, initial search distribution $\mu \in R^d$, $A = \sigma \mathbb{I} \in R^{d \times d}$, source distribution $\mu_\varphi \in R^d$, $A_\varphi \in R^{d \times d}$, initial mixture coefficient $\alpha_\varphi \in [0,1]$
$\alpha_\pi \leftarrow 1 - \alpha_\varphi$
**while** *stopping criteria not met* **do**
    $p \leftarrow \alpha_\varphi$ **if** *transfer activated* **else** $p \leftarrow 0$
    **for** $k \in \{1, 2, \dots, \lambda\}$ **do**
        sample $z_k \sim N(0, \mathbb{I})$, $s_k \sim bernoulli(p)$
        **if** $s_k = 1$ **then**
            $w_k \leftarrow A_\varphi z_k + \mu_\varphi$
            $d_k \leftarrow w_k - \mu$
            $\widetilde{w}_k \leftarrow \mu + d_k \times \min\left(1, \frac{r}{\|A^{-1}d_k\|}\right)$
            $z_k \leftarrow A^{-1}\left(\widetilde{w}_k - \mu\right)$
        **else**
            $w_k \leftarrow A z_k + \mu$
        **end**
    **end**
    sort $\{(z_k, w_k)\}$ *w.r.t.* $f(w_k)$
    compute $\{u_k\}$
    $\nabla_\mu \mathcal{J} \leftarrow \sum_{k=1}^{\lambda} u_k\ z_k$
    $\nabla_A \mathcal{J} \leftarrow \sum_{k=1}^{\lambda} u_k\ (z_k z_k^T - \mathbb{I})$
    $\mu \leftarrow \mu + \eta_\mu \cdot \nabla_\mu \mathcal{J}$
    $A \leftarrow A \cdot \exp(1/2 \cdot \eta_A \cdot \nabla_A \mathcal{J})$
    **if** *transfer activated* **then**
        $\nabla_{\alpha_\pi} \mathcal{J}^m = \sum_{k=1}^{\lambda} u_k\ \frac{\pi(w_k)}{\alpha_\pi \pi(w_k) + \alpha_\varphi \varphi(w_k)}$
        $\nabla_{\alpha_\varphi} \mathcal{J}^m = \sum_{k=1}^{\lambda} u_k\ \frac{\varphi(w_k)}{\alpha_\pi \pi(w_k) + \alpha_\varphi \varphi(w_k)}$
        $\alpha_\pi \leftarrow \alpha_\pi + \eta_\alpha \cdot \nabla_{\alpha_\pi} \mathcal{J}^m$
        $\alpha_\varphi \leftarrow \alpha_\varphi + \eta_\alpha \cdot \nabla_{\alpha_\varphi} \mathcal{J}^m$
        normalize $(\alpha_\pi, \alpha_\varphi)$ *s.t.* $\alpha_\pi + \alpha_\varphi = 1$
    **end**
**end**

---

TABLE I: TUNING PARAMETERS OF tNES

| |
|---|
| $\{\Delta t, t_{max}\}$: transfer plan |
| $\lambda$ : population size |
| $r$: Mahalanobis distance threshold for pseudo-offspring projection |
| $u_k$: utility function |
| $lr = \{\eta_\mu, \eta_A, \eta_\alpha\}$ : learning rate |

target distribution and follow the original xNES procedure. Otherwise, pseudo-offspring are sampled from the mixture model (6) and their fitness are evaluated. Additionally, projection (10) is applied to all pseudo-offspring from the source distribution, after which a coordinate transformation $z_k = A^{-1}\left(\widetilde{w}_k - \mu\right)$ maps the projected pseudo-offspring to the natural coordinates of the target distribution (this step is required for xNES updates). Then, fitness shaping is applied. All pseudo-offspring (now $z_k$) are sorted by their fitness, i.e., $f(w_1) \geq \cdots \geq f(w_\lambda)$, and the rank-based utility value:

$$u_k = \frac{\max(0, \log(\frac{\lambda}{2}+1) - \log k)}{\sum_{j=1}^{\lambda} \max(0, \log(\frac{\lambda}{2}+1) - \log j)}, \tag{11}$$

TABLE II: THE PHYSIC-INFORMED NEURAL NETWORK AND OPTIMIZATION CONFIGURATIONS USED IN EXPERIMENTAL STUDY

| Example | A<br>1D steady state<br>convection-diffusion | B<br>2D projectile motion | C<br>Model equations of traveling waves | |
|---|---|---|---|---|
| | | | (1) Linearized Burgers<br>(2) Nonlinear Burgers | (3) Korteweg–de Vries<br>(KdV) |
| PINN architecture | $(x) - 5 - 5 - (\hat{T})$ | $(t) - 3 - 3 - 3 - 3 - (\hat{x}, \hat{y})$ | $(x, t) - 4 - 4 - 4 - (\hat{u})$ | $(x, t) - 4 - 4 - 4 - 4 - (\hat{u})$ |
| dim($\boldsymbol{w}$) | 45 | 45 | 56 | 72 |
| no. collocation points $m$ sampled for<br>1 evaluation of loss $\mathcal{L}_{PDE} + \mathcal{L}_{BC/IC}$ | $1000 + 2$ | $1000 + 1$ | $5000 + 50$ | $10000 + 100$ |
| *t*NES<br>&<br>xNES<br>setting | max. evaluation | 2e5 | 2e5 | 2e5 | 3e5 |
| | population size $\lambda$ | 20 | 20 | 20 | 30 |
| | learning rate $lr =$<br>$\{\eta_\delta, \eta_M, \eta_\alpha\}$ | 1, 5e-2, 5e-2 | 1, 5e-2, 5e-2 | 1, 1e-2, 1e-2 | 1, 1e-2, 1e-2 |
| | initial search<br>distribution $\{\mu, \sigma \mathbb{I}\}$ | 0, 5e-2 | 0, 5e-2 | 0, 5e-2 | 0, 5e-2 |
| *t*NES transfer plan: $\{\Delta t, t_{max}\}$ | | 2, 500 | 2, 500 | 2, 500 | 2, 1000 |
| SGD<br>(ADAM)<br>setting | max. evaluation | 2e5 | 2e5 | 2e5 | 3e5 |
| | initial learning rate | 5e-2 | 5e-2 | 1e-2 | 1e-2 |
| | learning plan | reduce learning rate by half on plateau, with a min. learning rate set at 1e-6 | | | |

Note: the *t*NES, &, xNES setting rows span across A, B, C(1)(2), C(3) columns.

(1) For the PINN architecture, the numbers in between input and output represent the number of nodes in hidden layers. For example, $(t) - 3 - 3 - 3 - 3 - (\hat{x}, \hat{y})$ indicates a neural network with single input $t$, followed by 4 hidden layers with 3 nodes in each layer, and multi-output $(\hat{x}, \hat{y})$. All hidden layers, except the final hidden layer, include a bias term and use 'tanh' activation function. The final hidden layer uses 'linear' activation function and does not include a bias term.

(2) In *t*NES, $r = \sqrt{12}$ is used as the Mahalanobis distance threshold for pseudo-offspring projection.

(3) For neural network optimization, in additional to the maximum number of evaluations, a target loss = 1e-9 is also set as stopping criteria.

(4) Keras package is used to perform ADAM. For ADAM parameters except those stated above, default values are used.

is computed to replace the $k$-th best fitness $f(\boldsymbol{w}_k)$. Subsequently, the algorithm follows the xNES procedure for the computation of gradient estimates and search distribution update. The *t*NES algorithm also updates the mixture model coefficients as per (9) in the transfer mode, before entering the next iteration. The optimization loop terminates once the stopping criteria, for example, a target loss or maximum iterations, is met. The algorithm outputs the best found pseudo-offspring (solution) for the target problem as well as the optimized search distribution, which is added to the solved problems library as an experiential prior for future use.

There are some tunable parameters in the *t*NES algorithm; see TABLE I. The predetermined transfer plan schedules the transfer to take place after every $\Delta t$ iterations, and until $t_{max}$ iterations. Although it is safe to specify a large $t_{max}$ since the adaptive design will auto-terminate the transfer once it is no longer beneficial, setting an appropriate $t_{max}$ can prevent unnecessary computations. Other tunable parameters include the pseudo-offspring projection threshold, population size, learning rates and utility function. Like its baseline xNES algorithm, *t*NES is sensitive to these parameters, and the best settings may vary across problems.

## V. EXPERIMENTAL STUDY

This section contains empirical demonstrations that neuroevolution and transfer neuroevolution are noteworthy approaches for solving differential equations. Results on several pedagogical examples representative of real world phenomena: *A. 1D steady state convection-diffusion equation*, *B. 2D projectile motion equations*, and *C. model equations of traveling waves*, are presented. In each example, some parameter in the differential equation or initial condition is subject to change, forming multiple related problems to be solved. We investigate physics-informed neural networks to emulate the solution to the differential equations. The neural network architecture is fixed beforehand, without carrying out exhaustive architecture search. The selected architecture is nevertheless robust enough to produce good approximations for different problems derived from the same differential equation example. By altering the neural network weights, the differential equation as well as the prescribed initial and/or boundary conditions must be satisfied. The resultant global optimization problem can often be very challenging even if the differential equation looks simple on the surface.

The configurations of the physics-informed neural networks and the optimization algorithms for different examples are given in TABLE II. We compare *t*NES (instantiation of transfer neuroevolution) and xNES (example of neuroevolution) with the ADAM (state-of-the-art variant of SGD) algorithm [51]. A similar setting for all 3 optimizers is used for fairness of comparisons. For each loss (fitness) evaluation, the sum of mean squared residuals from the differential equations as well as the initial and boundary conditions (2) is computed over $m$ randomly sampled collocation points. Similar to the network architecture, we do not exhaustively search for the best optimization setting. A sufficiently robust setting that works across different problems derived from the same differential equation example is chosen. It is reckoned that the convergence of SGD (ADAM) is highly sensitive to the learning rate. Hence, a learning plan is adopted to reduce its learning rate by half on plateaus, until a minimum learning rate of 1e-6 is reached. This strategy significantly improves the robustness and performance

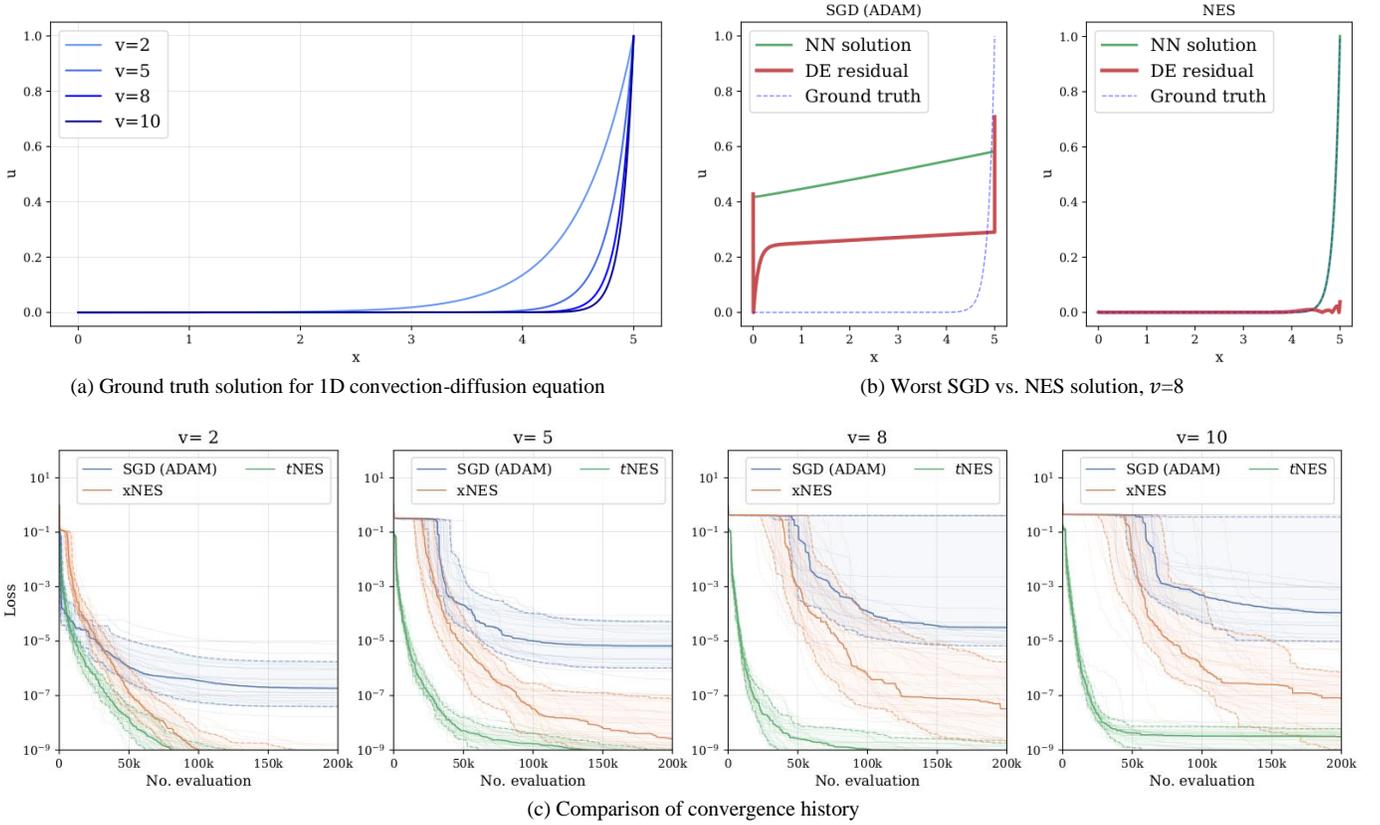

(a) Ground truth solution for 1D convection-diffusion equation

(b) Worst SGD vs. NES solution, $v$=8

(c) Comparison of convergence history

Fig. 5. (a) Ground truth (analytical) solution for 1D steady state convection-diffusion equation, with $v$=2, 5, 8, 10 in (12). The underlying optimization problem is getting more challenging when the velocity $v$ increases. For example, at $v$=8, the SGD solution shown in (b) is not meaningful because it could not fulfill the boundary conditions, thus it induces high residual at the two boundaries of $x$. The worst of xNES is seen to perform notably better than SGD. In (c), the convergence trends from 30 independent SGD, xNES and $t$NES optimization runs are plotted. The bold lines indicate the mean convergence path, and the shaded areas indicate the interpercentile range from their 10th-90th percentiles. In all problems, the difference in performance (at 200k evaluations) between three optimization algorithms is concluded by the Friedman tests at 5% significance level. Both xNES and $t$NES perform better than SGD at 200k evaluations; furthermore, $t$NES significantly outperforms xNES when considered under a limited computational budget of 50k evaluations (concluded by Mann-Whitney rank test, at 5% significance level).

of SGD by preventing it from being easily trapped in a plateau and also speeding up the convergence.

### A. 1D Steady State Convection-Diffusion Equation

The convection-diffusion equation describes a very common phenomenon where a physical quantity, such as particles, energy, and temperature, are transferred inside a physical system due to two processes: convection and diffusion (also known as directional and non-directional transfer). As a motivating example, a simplified 1D steady state convection-diffusion equation is solved:

$$v \cdot u_x = k \cdot u_{xx}, \ x \epsilon [0, L], \quad (12)$$

with the boundary conditions $u$=0, $x$=0; $u$=1, $x = L$. We are interested in the solution $u(x)$, the temperature in the spatial domain $x \epsilon [0, L]$. The velocity $v$, diffusion coefficient $k$, and domain boundary $L$ are the problem specific constants. In this special case, the ground truth solution can be analytically derived as:

$$u(x) = \frac{1 - \exp(x \cdot v / k)}{1 - \exp(L \cdot v / k)}, \quad (13)$$

which is used to verify our optimized PINN solution. In our example, a domain boundary of $L$=5 and diffusion coefficient of $k$=1 is used. We solve the equation for different velocity values, $v$=2, 5, 8, 10. Classical numerical schemes may be able to solve this differential equation faster, but are notoriously prone to failure at high velocity values caused by inappropriate meshing, which gives rise to unphysical oscillations in the solution [52]. Being mesh-free, the PINN approach avoids this issue.

A physics-informed neural network with 45 tunable weights is constructed to emulate the solution $u$ given input $x$. Specifically, the loss function for the 1D steady state convection-diffusion equation (12) is defined as follows:

$$\mathcal{L} = \mathcal{L}_{DE} + \mathcal{L}_{BC}$$
$$= \frac{1}{m} \sum_{i=1}^{m} (k \cdot \hat{u}_{xx} - v \cdot \hat{u}_x)^2 + [\hat{u}(0) - 0 + \hat{u}(L) - 1]. \quad (14)$$

Note that in this example, the loss function (14) computes the mean squared residual of the differential equation over $m$=1000+2 collocation points, i.e., 1000 randomly sampled points within domain $x \epsilon [0, 5]$ using Latin hypercube sampling, plus the 2 boundary points $x$=0, $x = L$. In addition, the 2 boundary points are matched against the boundary conditions. We refer to such computation as a single evaluation of the

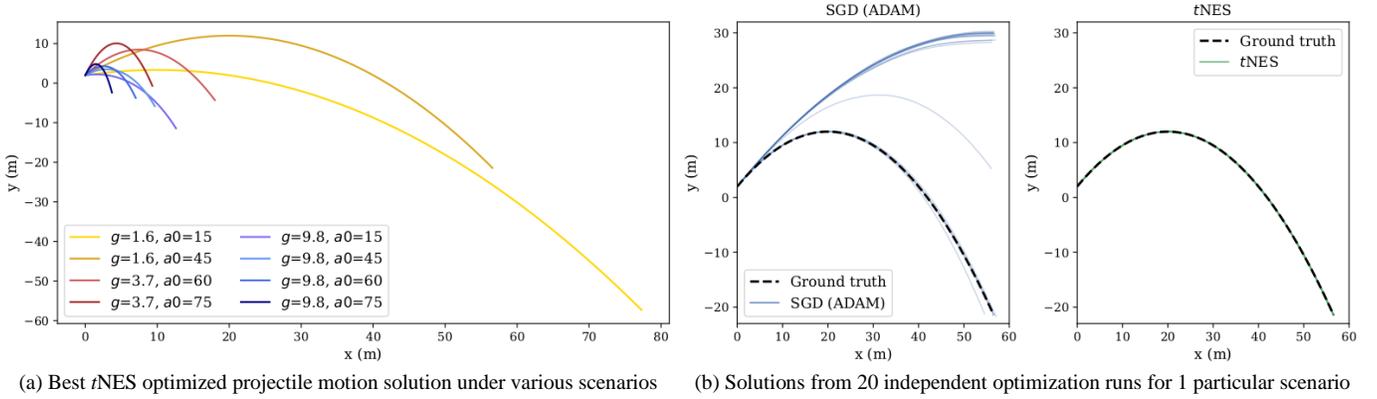

(a) Best $t$NES optimized projectile motion solution under various scenarios

(b) Solutions from 20 independent optimization runs for 1 particular scenario

Fig. 6. (a) The projectiles under the effect of Earth ($g$=9.8, $\rho$ = 1.2), Mars ($g$=3.7, no drag effect) and Moon ($g$=1.6, no drag effect) are respectively solved for $t$=0-2s, $t$=0-4.5s and $t$=0-10s. (b) All individual solutions given by 20 SGD and 40 $t$NES independent optimization runs for one particular scenario ($g$=1.6, $a0$=45, $t$=0-10s) are plotted and compared against the ground truth. The SGD runs lead to three different groups of projectiles, and deviations are observed within the two main groups. By knowledge transfer from a relevant prior (source distribution obtained by solving the $g$=3.7 at the same $a0$), all $t$NES runs result in an indistinguishable projectile that overlaps almost exactly with the ground truth.

neural network loss. Since $t$NES, xNES and SGD algorithms are all stochastic in nature, multiple optimization runs are performed for each problem to statistically assess their results.

The results are summarized in Figure 5. At larger $v$ value, the solution $u(x)$ is characterized by an extended flat region followed by steep gradient curvature near the end of $x$ domain (Figure 5a). As the nonlinearity of the solution increases with velocity $v$, the underlying differential equation is more difficult to match. This makes the optimization problem become increasingly challenging. As a consequence, some of the solutions returned by SGD optimization are not meaningful, because they could not even fulfill the boundary conditions (Figure 5b). These poorly optimized solutions have relatively high residual terms from the boundary conditions at two ends of $x$, and from the differential equation over the entire domain. In contrast, the worst solution from neuroevoluion performs much better than SGD in terms of the optimized loss (aggregated residual from the differential equation and boundary conditions) and the error against ground truth solution. For example, at $v$=8, the aggregated residuals from the worst xNES and SGD solution are 1.3e-5 $vs.$ 4.1e-1, and their mean squared errors against ground truth are 2.3e-7 $vs.$ 2.3e-1.

Figure 5c compares the convergence trends of different optimizers, derived from 30 independent runs for each problem. We first compare the xNES with SGD. At lower $v$, SGD descends very quickly, but starts levelling off even though the loss values are far from the optimum. In contrast, xNES starts to catch up with SGD given a sufficient number of evaluations, and it eventually converges (at least 2 orders of magnitude) better. As the problem gets more difficult when $v$ increases, both xNES and SGD spend a significant amount of early evaluations on a plateau before they can find a right path to descend. This is when the SGD fails to find a good solution, because on some occasions they could not find a right path to descend from the early plateau. Overall, xNES converges better than SGD in both speed and accuracy when $v$ increases. This simple example demonstrates the promise of neuroevolution in solving problems where SGD gets trapped. We further demonstrate the benefit of transfer neuroevolution in Figure 5c.

For each problem, source distributions were obtained by solving the same equation with $v - 5$ using xNES; here, $v$ is the velocity in the target problem. Then 30 independent $t$NES runs are performed. The proposed transfer method successfully speeds up the convergence (quickly escapes from the plateau) and also helps to achieve a better optimized solution by exploiting experiential priors. In this example, the optimized loss achieved by neuroevolution ($t$NES & xNES) are at least 2 orders of magnitude lower than SGD. Furthermore, under a limited computational budget of 50k evaluations, $t$NES significantly outperforms xNES (as per the Mann-Whitney rank test) across all the problems.

### B. 2D Projectile Motion Equations

In our second example, transfer neuroevolution is applied to solve the physics of projectile motion. Assuming a ball is thrown into the air at specific launch angle $a_0$, initial velocity $vel_0$ and location $(x_0, y_0)$, the entire projectile of the flying ball can be predicted by solving the following ordinary differential equations:

$$x_{tt} = -R \cdot x_t \qquad, \quad t\epsilon[0, T], \tag{15a}$$
$$y_{tt} = -R \cdot y_t - g, \quad t\epsilon[0, T], \tag{15b}$$

subject to the initial conditions $x = x_0$, $x_t = vel_0 cos(a_0\pi/180)$, $t$=0, and $y = y_0$, $y_t = vel_0 sin(a_0\pi/180)$, $t$=0. The solution to the above differential equations gives the horizontal and vertical position $x(t)$ and $y(t)$ of the flying ball, within the time domain $t\epsilon[0, T]$. Now, let us assume the following initial conditions: $(x_0, y_0)$=(0, 2m), $vel_0$=8m/s, i.e., the ball is released at 2m height, at an initial velocity of 8m/s (simulating a human throwing a basketball-sized ball with his arm). We vary the launch angles to simulate different projectiles. In (15), $R$ is the resistance coefficient which is related to the air density and object velocity, and $g$ is the gravity. For simplicity, it is assumed that the ball does not spin much so the Magnus effect is neglected. Then the air resistance or drag effect is modelled by $R = CV$, where $V = \sqrt{(x_t^2 + y_t^2)}$, $C = 0.5\rho C_d A/m$. $\rho$ is the air density. $C_d$ is the drag coefficient, $A = \pi r^2$ is the

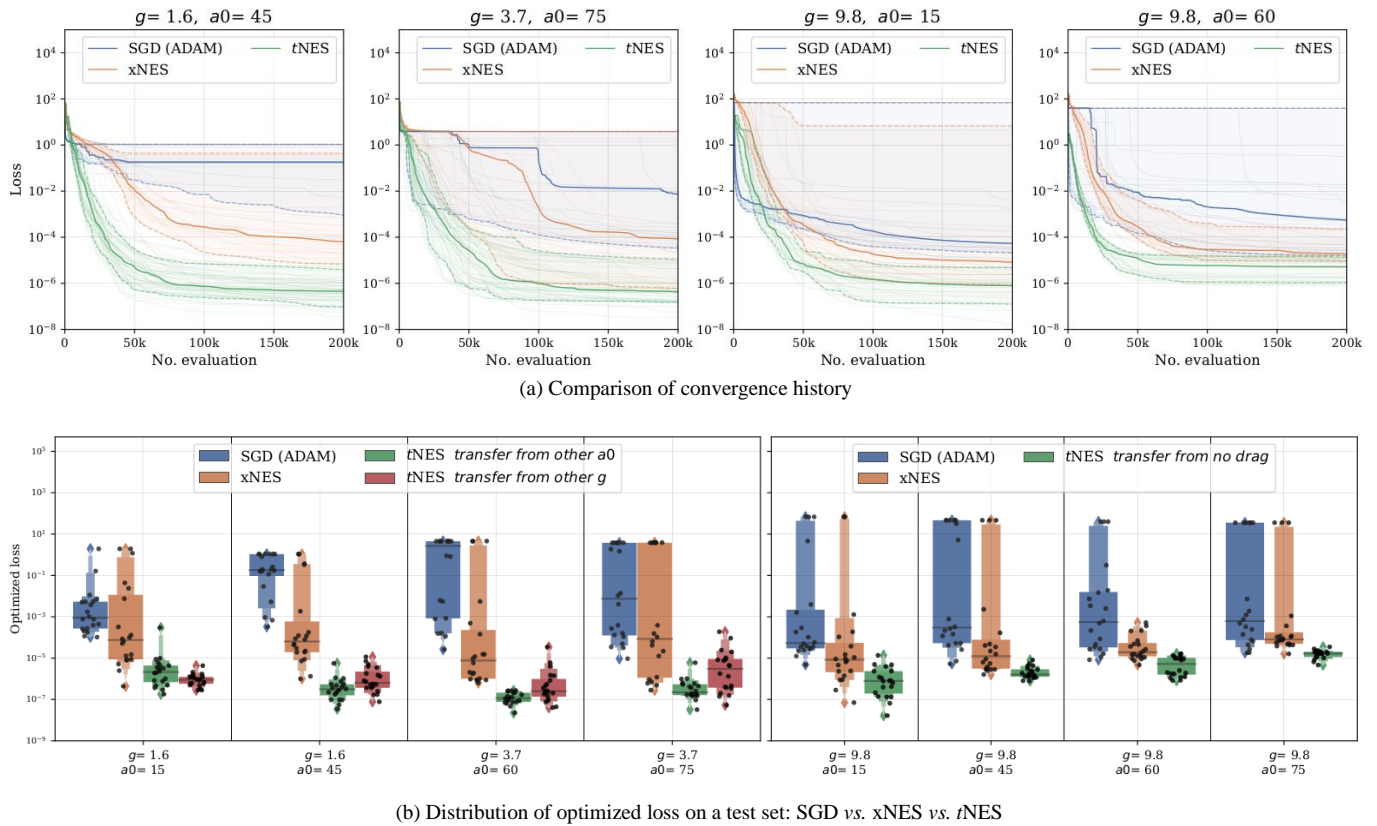

(a) Comparison of convergence history

(b) Distribution of optimized loss on a test set: SGD *vs.* xNES *vs.* *t*NES

Fig. 7. (a) The convergence trends from 20 independent SGD and xNES optimization runs, and 40 *t*NES optimization runs for selected scenarios are displayed. The bold lines indicate the mean convergence path and the shaded areas indicate their $10^{th}$-$90^{th}$ intercentile ranges. (b) The distribution of optimized loss on a test set given by tNES, xNES and SGD for all scenarios are compared. Some scenarios (e.g., $g$=1.6, $a0$=45 & $g$=3.7, $a0$=60) are challenging to SGD but not to the neuroevolution (xNES & *t*NES). For both $g$=1.6 and $g$=3.7, the results (in green blocks) given source problem from the same $g$ (i.e., from $a0$=15 to $a0$=30, from $a0$=60 to $a0$=75, and vice versa), and the results (in red blocks) given source problem from the same $a0$ (i.e., from $g$=3.7 to $g$=1.6, and from $g$=9.8 to $g$=3.7), are compared. For $g$=9.8, the source problem came from the same scenario without considering the drag effect. Comparing their optimized loss on a test set, the Friedman tests conclude a difference in performance between three optimization algorithms in all scenarios (at 5% significance level). In particular, xNES performs better than SGD in all scenarios except for $g$=3.7, $a0$=75, whereas *t*NES performs significantly better than both SGD and xNES (as per the Mann-Whitney rank test at 5% significance level).

cross-sectional area of the ball, and $m$ is its mass. The radius and mass of a basketball are typically $r$= 0.12 (m) and $m$=0.6 (kg), and the drag coefficient $C_d$ is around 0.54 [53].

Figure 6a shows the *t*NES optimized projectiles of the flying ball for various launch angles $a_0$=15, 30, 45, 75, between the time domain $t\epsilon[0, 2s]$ under the effect of Earth gravity $g$=9.8 (m/s²) and air $\rho$ = 1.2 (kg/m³). Moreover, *t*NES is applied to solve the projectiles on another planet such as Mars ($g$=3.7, $t$=0-4.5s, no drag effect) or on the Moon ($g$=1.6, $t$=0-10s, no drag effect), since the same physics laws apply elsewhere in the universe. In Figure 6b, solutions from 20 SGD and 40 *t*NES optimization runs for one particular scenario ($g$=1.6, $a0$=45) are compared against the ground truth. The SGD produces three different groups of projectiles, two of which are poorly optimized solutions due to being trapped in some bad local minima. There are also deviations within the groups, because the optimization runs are yet to fully converge into the minima. In contrast, the quality of *t*NES solutions are consistent, so they all overlap onto an indistinguishable projectile. Their median mean squared errors against the ground truth solution are 2.6e2 for SGD and 6.9e-7 for *t*NES. A similar pattern can be observed in other scenarios as well.

The convergence and optimized loss results given by *t*NES, xNES and SGD are summarized in Figure 7. We show the convergence history for selected scenarios, and compare the optimized loss for all scenarios with a "letter value" plot [54]. Different from previous examples, for most of the problems here the xNES cannot completely avoid being trapped in a bad local minimum. Nevertheless, it offers a better chance to find better minima when compared to SGD. At the same time, *t*NES gives the best results by leveraging the experiential priors. These priors come from solving a related problem scenario in the absence of drag, a scenario with different gravitational conditions but a similar launch angle, or for one with different launch angle but similar gravity. By transferring and reusing information from such relevant sources, *t*NES avoids being trapped in a bad local minimum. These results demonstrate that transfer neuroevolution indeed solves differential equations faster and better.

### C. Model Equations of Traveling Waves

In this section, transfer neuroevolution is applied to solve for several transient PDEs which describe traveling wave(s). The first model equation—Burgers' equation—can be viewed as a simplified 1D version of the Navier–Stokes equations, for

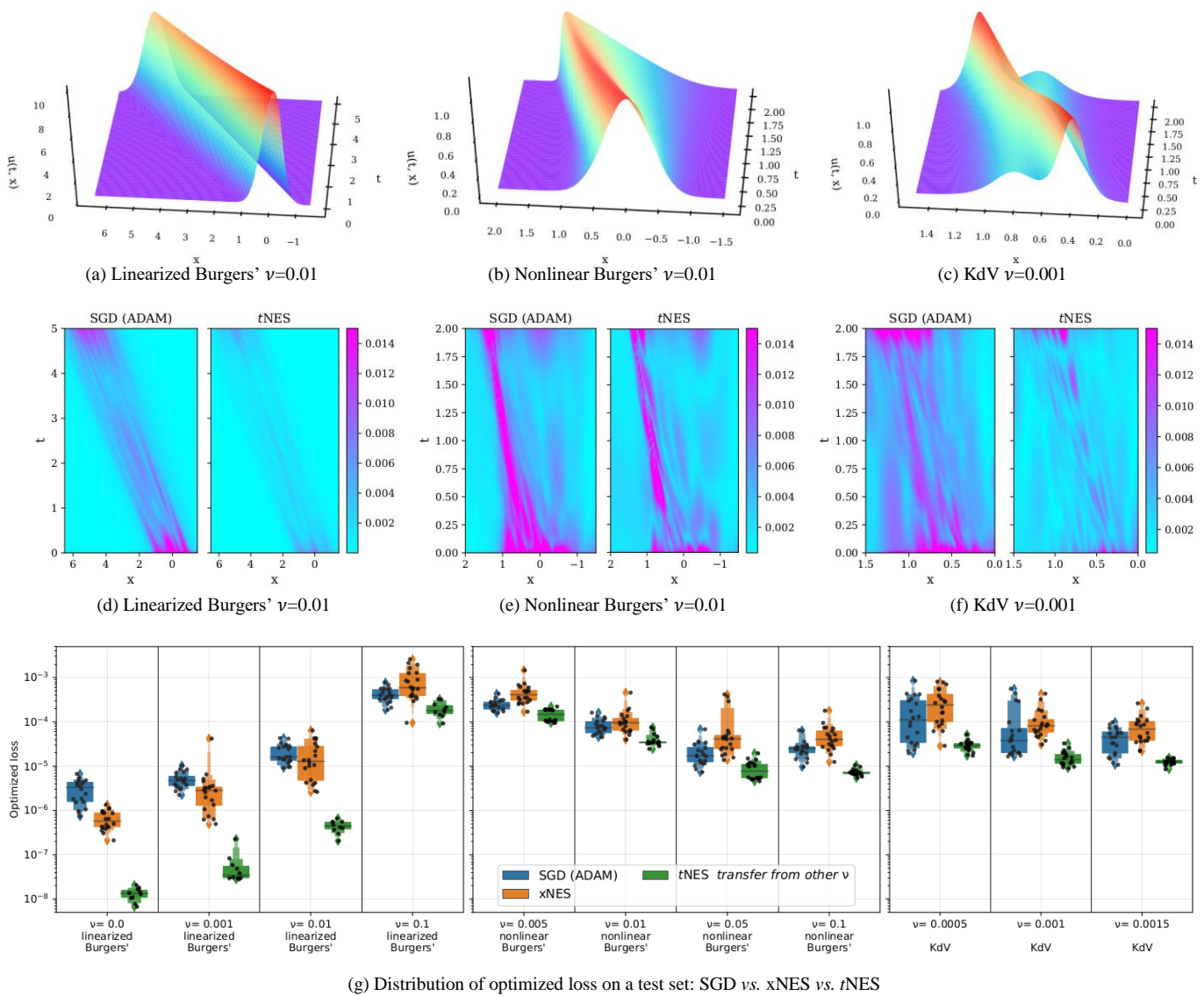

(a) Linearized Burgers' $\nu$=0.01

(b) Nonlinear Burgers' $\nu$=0.01

(c) KdV $\nu$=0.001

(d) Linearized Burgers' $\nu$=0.01

(e) Nonlinear Burgers' $\nu$=0.01

(f) KdV $\nu$=0.001

(g) Distribution of optimized loss on a test set: SGD *vs.* xNES *vs.* *t*NES

Fig. 8. (a)-(c) Sample solution to transient PDE problems generated by *t*NES. (d)-(f) Mean absolute residual maps (aggregated from both differential equation and initial condition) from 20 SGD and *t*NES runs. They are evaluated on a test set of 10k uniform grid points across respective problem domain. Results show that *t*NES performs better than SGD at regions near initial state at *t*=0 and also regions with steep gradients for all 3 problems. (g) The distributions of optimized loss on a test set given by *t*NES, xNES and SGD are compared for all transient PDE problems. For the linearized Burgers' equation problems, the same source problem $\nu$=0.02 is used by the *t*NES. For the nonlinear Burgers' equation problems $\nu$=0.005 & 0.01, source problems came from $\nu$=0.006 & 0.012; for problems $\nu$=0.05 & 0.1, source problems came from $\nu$=0.06 & 0.12. For the KdV equation problem $\nu$=0.0005, source problems came from $\nu$=0.0004 & 0.0008; for problem $\nu$=0.0010, source problems came from $\nu$=0.0008 & 0.0016; for problem $\nu$=0.0015, source problems came from $\nu$=0.0012 & 0.0016. The Friedman tests conclude a difference in performance between three optimization algorithms at 5% significance level. In particular, the *t*NES outperforms xNES and SGD (ADAM) at 5% significance level (as per the Mann-Whitney rank test) across all the problems considered in this example.

understanding the dynamics of fluids and other important physics [55]. The Burgers' equation has a general form:

$$u_t + u \cdot u_x = \nu \cdot u_{xx}, \qquad (16)$$

which combines the nonlinear advection and diffusion effects. The diffusion coefficient (or kinematic viscosity) $\nu$ is an important parameter for characterizing the fluid properties.

### 1) Linearized Burgers' Equation

We firstly demonstrate the efficacy of transfer neuroevolution on a simpler version, linearized Burgers' equation (it can be viewed as a transient extension of our first

example's 1D steady state convection-diffusion equation):

$$u_t + c \cdot u_x = \nu \cdot u_{xx}, \quad x \epsilon [-1.5, 6.5], \quad t \epsilon [0, 5], \qquad (17)$$

with an initial condition $u(x, 0) = 10 \cdot \exp(-(2 \cdot x)^2)$. In this setup, a single waveform is propagated at constant velocity $c$=1. We vary the diffusion coefficient $\nu$=0, 0.001, 0.01, 0.1 to form different problems for transfer neuroevolution to solve; in all cases, the experiential prior is drawn from the optimized search distribution for $\nu$=0.02. At $\nu$=0, there is no diffusion effect so that the initial waveform maintains its form, while at larger $\nu$ the waveform loses its magnitude over time due to the diffusion effect.

Figure 8a presents a *t*NES solution for the case $\nu$=0.01 (with $\nu$ =0.02 as source prior), where the propagation of the wave can be seen. In Figure 8d, the mean absolute residual (aggregated residual from differential equation and initial condition) for this case based on 20 *t*NES and SGD optimization runs are compared. On average, SGD gives higher residual than *t*NES near the diffusive wave region. Higher SGD residual is also observed in capturing the initial condition. The distributions of optimized loss given by *t*NES, xNES and SGD are compared in Figure 8g. The optimized loss values given by SGD and xNES are in a similar range. By leveraging the experiential priors, *t*NES gives (1-2 orders of magnitude) better optimized loss than SGD and xNES.

### 2) Nonlinear Burgers' Equation

Next, the full nonlinear Burgers' equation is considered:

$$u_t + u \cdot u_x = \nu \cdot u_{xx}, \ \ x\epsilon[-1.5, 2], \ \ t\epsilon[0, 2] \quad (18)$$

with an initial condition $u(x, 0) = \exp(-(2 \cdot x)^2)$. Figure 8b shows a *t*NES optimized solution for $\nu$=0.01 (with $\nu$ =0.006 as source prior), illustrating the nonlinear waveform propagation with both compression and rarefaction effects. A steep gradient can be observed on one side. In Figure 8e, the mean absolute residual (aggregated residual from differential equation and initial condition) based on 20 independent *t*NES and SGD optimization runs for the same problem are compared. The overall results show that *t*NES performs better than SGD at the initial stages (near $t$=0) and near the steep gradient region, suggesting a more accurate solution from *t*NES. The results from multiple optimization runs for problems with $\nu$=0.005, 0.01, 0.05, 0.1 are compared in Figure 8g. The optimized loss values given by both xNES and SGD are around 1e-3 to 1e-5 and could not be further reduced. This is mainly caused by the neural networks' inability to emulate the right solution at steep gradient region. Transferred from other diffusion coefficients, *t*NES achieves the best optimized loss among all 3 techniques.

### 3) Korteweg–de Vries (KdV) Equation

The KdV equation [56] is used in physics and engineering to model the weakly nonlinear long waves (for example, waves on shallow water surfaces). It has several variants, and we consider the KdV equation in the form:

$$u_t + u \cdot u_x = \nu \cdot u_{xxx}, \ \ x\epsilon[0, 1.5], \ \ t\epsilon[0, 2] \quad (19)$$

which consists of a dispersive coefficient $\nu$. The solution of (19) describes the height of the wave at position $x$ and time $t$. We consider the following initial condition [57]: $(x, 0) = 3c_1 \mathrm{sech}^2 a_1(x - x_1) + 3c_2 \mathrm{sech}^2 a_2(x - x_2)$. Specifically, $c_1$=0.3, $c_2$=0.1, $x_1$=0.4, $x_2$=0.8, $a_1$=$0.5\sqrt{c_1/\nu}$, and $a_2$=$0.5\sqrt{c_2/\nu}$. With this initial condition, the equation simulates the collision of 2 waves of different magnitudes traveling from different locations. These 2 waves gradually lose their magnitudes while they approach each other, and then collide. After the collision, they split and regain their magnitudes. Figure 8c shows such interaction for $\nu$ =0.001 using the solution optimized by *t*NES (with $\nu$ =0.0008 as source prior). Figure 8f compares the mean absolute residual (aggregated residual from differential equation and initial condition) between *t*NES and SGD solutions for the same case, based on 20 independent optimization runs. The results suggest a more accurate solution from *t*NES across the computational domain. Figure 8g shows the distribution of the optimized loss based on multiple *t*NES, xNES and SGD optimization runs for dispersive coefficient $\nu$=0.0005, 0.0010, 0.0015. Transferred from other diffusion coefficients, *t*NES improves the optimized loss significantly in comparison to both SGD and xNES.

### D. Study of Mixing Coefficients in tNES

In this section, we provide visualization and discussion on how the mixing coefficients change during the evolutionary process, under the proposed mixture model-based adaptive transfer. Taking the 2D projectile motion as example, the target problem is to solve the projectile motion under the effect of moon gravity $g$=1.6, with an initial launch angle $a0$=45.

Firstly, a multi-source setup is considered, where one of the sources is less relevant (optimized search distribution for $g$=9.8, $a0$=45) to the target than the other (optimized search distribution for $g$=3.7, $a0$=45). By default, the mixture model for transfer neuroevolution is initialized with equal mixing coefficients for the target and all source distribution components. As shown in the left panel of Figure 9a, the mixing coefficient of the beneficial source quickly ascends to 1 at the early evolutionary stage, while suppressing the mixing coefficients from the less relevant source. This shows that the mixture model-based transfer is capable of jointly processing multiple sources and selecting the one that is most relevant to the target. After a certain point however, the source no longer remains competitive with the evolved target search distribution, which causes its mixing coefficient to gradually drop. Eventually, the target search distribution takes over the mixture model and all the source components are deactivated. Due to the directional bias induced by the source priors, the search progresses at a faster rate towards a better solution.

Next, the sensitivity of initial mixing coefficient αs' is investigated under the single source setup. We consider the scenario where the single source is from a related prior (optimized search distribution for $g$=3.7, $a0$=45), and is initialized at different levels of $\alpha_\varphi$ = 0.99, 0.1, 0. As shown in the right panel of Figure 9a, $\alpha_\varphi$ quickly ascends to dominate the mixture model at the early evolutionary stage, even when it has a small initial value. That is, the proposed mixture model-based adaptive transfer method is not very sensitive to the initial αs' setting. The only exception is when the initial $\alpha_\varphi$ is set to 0, since in this case pseudo-offspring will never be drawn from the source distribution. Thus, transfer never occurs. On the other hand, if the single source forms an unrelated prior (far from the optimum target distribution), its mixing coefficient quickly descends to 0 so that negative transfer is avoided. As a result, the search has a similar convergence trend as the no-transfer scenario (right panel of Figure 9b).

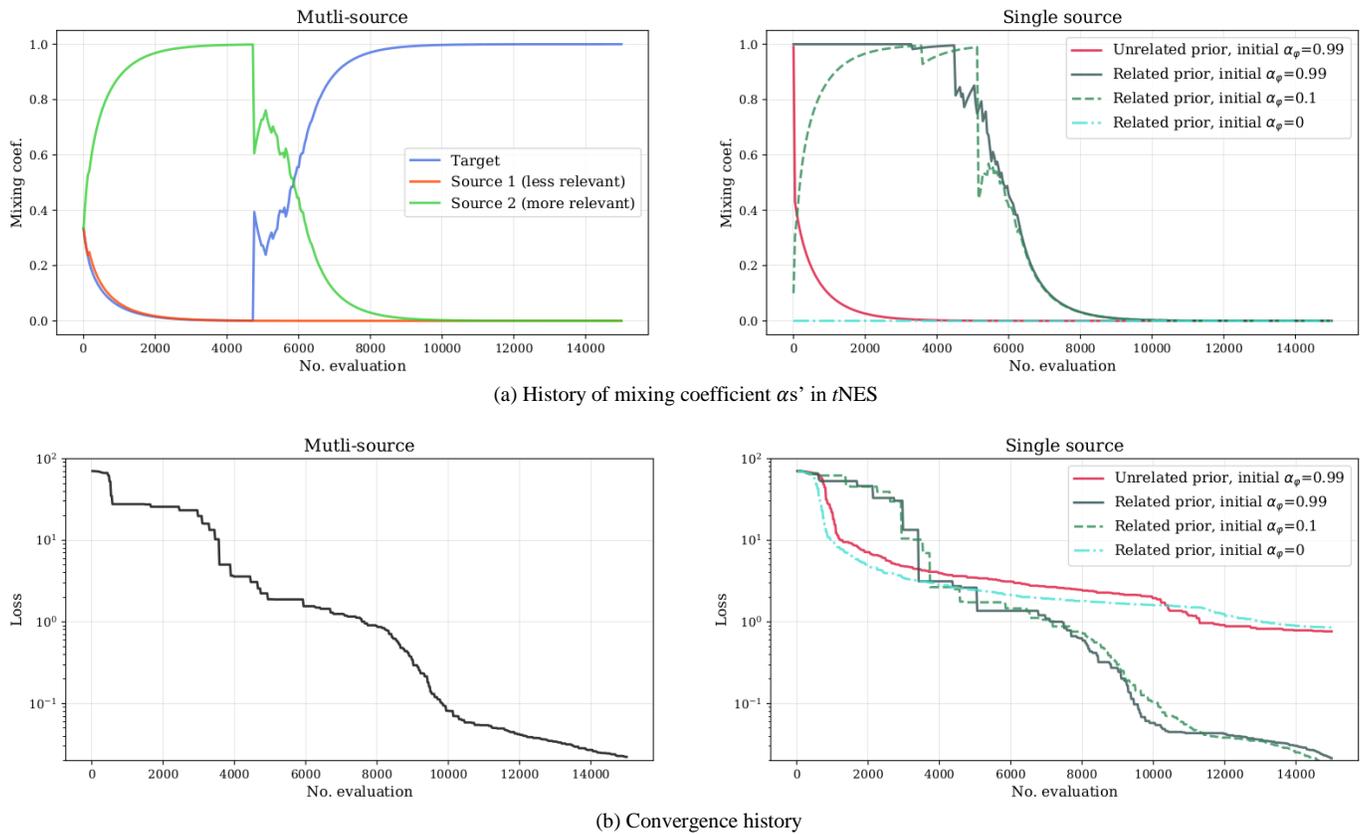

(a) History of mixing coefficient $\alpha$s' in $t$NES

(b) Convergence history

Fig. 9. (a) In the multi-source plot, the history of mixing coefficient $\alpha$s' from all mixture model components during a $t$NES search process are shown. In the single source plot, the history of mixing coefficient of a single source is shown under different initial values of $\alpha_\varphi$. Their respective convergence trends between 0 and 15,000 evaluations are shown in (b), which highlights that effective transfer neuroevolution leads to better convergence.

## VI. CONCLUSIONS

In conclusion, this paper demonstrated neuroevolution as a notable approach for solving differential equations, where the problem has been transformed to one of global optimization of physics-informed neural networks. In such problems, we merit the accuracy of the solution, which shall be produced by a better optimized neural network. Gradient descent methods such as SGD may not always be the best approach, because they are prone to premature convergence. Therefore, we introduce neuroevolution, which naturally adapts a parallel exploration of diverse solutions during the optimization search to overcome local optima and to achieve better optimized neural networks. We empirically demonstrated that an implementation of the state-of-the-art NES algorithm could outperform SGD in many differential equations to give more accurate solutions.

Moreover, a novel transfer neuroevolution method which is suitable for general probabilistic distribution-based evolution strategies is proposed. It features a mixture model-based adaptive transfer mechanism to systematically transfer useful knowledge from relevant source problems, while protecting the search against negative transfer. In practice, it is very common for many similar differential equation problems to occur under different environments and boundary conditions; as such, we expect transfer neuroevolution to shine in this domain. We empirically demonstrated that transfer neuroevolution may not only improve the convergence speed but also improve the solution accuracy in comparison to the baseline neuroevolution algorithm.

In our experimental study, neuroevolution nevertheless starts losing its competitive advantage to SGD (ADAM) for more complex differential equations. On this account, it is important for future research to further improve the effectiveness of neuroevolution in terms of: (1) handling much larger neural networks, such that the solution of complex differential equations can be accurately emulated; and (2) simultaneously searching for the best network architecture in addition to the neural network weights [58, 59, 60, 61]. Another interesting idea is the development of multi-objective, multi-task neuroevolution [62, 63] in the context of physics-informed neural networks. We expect these approaches to have a good synergy with the recently proposed domain decomposition strategy [64, 65], where separate neural networks are used to approximate different sub-domains of a differential equation.

## ACKNOWLEDGEMENT

Jian Cheng Wong is supported by the Institute of High Performance Computing (IHPC), A*STAR. Abhishek Gupta is supported in part by the A*STAR Cyber-Physical Production System (CPPS) – Towards Contextual and Intelligent Response Research Program, under the RIE2020 IAF-PP Grant A19C1a0018. Yew-Soon Ong is supported by the Data Science and Artificial Intelligence Research Center (DSAIR) and the School of Computer Science and Engineering at Nanyang Technological University.